\documentclass{article}
\usepackage{spconf,amsmath,epsfig}
\usepackage{multirow}
\usepackage{hyperref}

\let\OLDthebibliography\thebibliography
\renewcommand\thebibliography[1]{
  \OLDthebibliography{#1}
  \setlength{\parskip}{0pt}
  \setlength{\itemsep}{0pt plus 0.3ex}
}

\begin{document}\sloppy

% Example definitions.
% --------------------
\def\x{{\mathbf x}}
\def\L{{\cal L}}

% Title.
% ------
\title{iDAT: inverse Distillation Adapter-Tuning}

%
% Single address.
% ---------------
% \name{Anonymous ICME submission}
%Address and e-mail should NOT be added in the submission paper. They should be present only in the camera ready paper. 

\name{Jiacheng Ruan$^{1}$, Jingsheng Gao$^{1}$, Mingye Xie$^{1}$, Daize Dong$^{2}$, Suncheng Xiang$^{1}$, Ting Liu$^{1}$, Yuzhuo Fu$^{1,*}$ \thanks{\textsuperscript{$\dagger$}Corresponding Author.}}
\address{$^1$Shanghai Jiao Tong University$\quad$
$^2$Shanghai Artificial Intelligence Laboratory \\
{\tt\small jackchenruan@sjtu.edu.cn}
}

\maketitle

\begin{abstract}

Adapter-Tuning (AT) method involves freezing a pre-trained model and introducing trainable adapter modules to acquire downstream knowledge, thereby calibrating the model for better adaptation to downstream tasks. This paper proposes a distillation framework for the AT method instead of crafting a carefully designed adapter module, which aims to improve fine-tuning performance. For the first time, we explore the possibility of combining the AT method with knowledge distillation. Via statistical analysis, we observe significant differences in the knowledge acquisition between adapter modules of different models. Leveraging these differences, we propose a simple yet effective framework called inverse Distillation Adapter-Tuning (iDAT). Specifically, we designate the smaller model as the teacher and the larger model as the student. The two are jointly trained, and online knowledge distillation is applied to inject knowledge of different perspective to student model, and significantly enhance the fine-tuning performance on downstream tasks. Extensive experiments on the VTAB-1K benchmark with 19 image classification tasks demonstrate the effectiveness of iDAT. The results show that using existing AT method within our iDAT framework can further yield a 2.66\% performance gain, with only an additional 0.07M trainable parameters. Our approach compares favorably with state-of-the-arts without bells and whistles. Our code is available at \href{https://github.com/JCruan519/iDAT}{https://github.com/JCruan519/iDAT}.

\end{abstract}
\begin{keywords}
Adapter-Tuning, Knowledge distillation, Weak-to-Strong
\end{keywords}
\section{Introduction}
\label{sec:intro}

The pretraining-finetuning paradigm has achieved excellent performance on various tasks by pretraining models on large-scale datasets and fine-tuning them on downstream tasks. However, as model sizes increase, especially Transformer-based models \cite{vitsurvey}, full parameter fine-tuning (FPT) becomes impractical across different tasks. FPT requires training and storing the full set of model parameters for each task, which is storage-intensive and prone to overfitting due to the often limited amount of downstream data.

\begin{figure}[!t]
    \centering
    \includegraphics [width=0.48\textwidth] {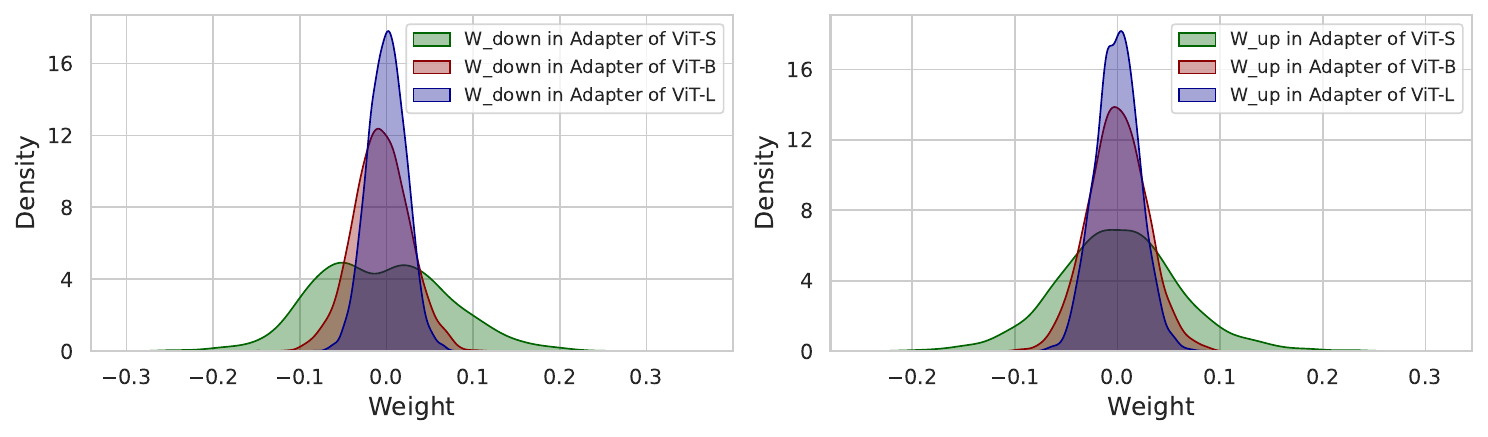}
    \caption{Weight distribution visualization on CIFAR-100 dataset. Our motivation: The differences in weight distributions signify the variances in the learned downstream knowledge \cite{parameterdistribution}. How to more effectively utilize these differences to enhance the model's fine-tuning performance?}
    \label{fig:motivation}
\end{figure}

Recently, the research community has focused on developing parameter-efficient fine-tuning methods across various domains \cite{nlpadapter, VPT, nlplora, LAMM}. These techniques, by freezing the pretrained backbone models and introducing a minimal number of learnable parameters for fine-tuning, have achieved performance surpassing that of FPT. Specifically, Adapter-Tuning (AT), a popular technique, incorporates learnable adapter modules (\textit{e.g.}, linear-activation-linear structures) into the frozen pretrained models to acquire downstream knowledge. Existing works often concentrate on introducing meticulously designed modules to attain a more comprehensive understanding of downstream knowledge. In contrast, beyond designing complex knowledge extraction modules, knowledge distillation can be employed as a simpler and more intuitive method for a more inclusive knowledge acquisition, which has been less explored.

In this paper, for the first time, we explore the potential of integrating the knowledge distillation framework with the AT method. Knowledge distillation involves the transfer of knowledge from a teacher model to a student model for improving the performance of student model. This transfer is depended on the presence of a knowledge difference. As illustrated in Figure \ref{fig:motivation}, we begin by visualizing the weights of the linear layers in adapter modules of different scales of ViTs \cite{vit} fine-tuned by Parallel Adapter \cite{adaptformer}, one of the classical AT methods\footnote{During downstream fine-tuning, the AT method updates the weights and biases of the linear layers in the adapter modules. This paper focuses primarily on these weights, omitting bias terms for simplicity.}. We observe that the weight distributions in adapter modules of ViT-B and ViT-L models are more concentrated, whereas those in the ViT-S model are more dispersed. This phenomenon reveals that the acquisition of downstream task knowledge by adapter modules differs according to the size of the model. Consequently, we investigate whether this inherent disparity can serve as additional knowledge to facilitate improved learning in models, thereby enhancing the performance of downstream fine-tuning.

Based on the observations and considerations outlined above, we propose the inverse Distillation Adapter-Tuning (iDAT) framework. iDAT harnesses the knowledge disparities across various adapters by employing a unique inverse knowledge distillation process, which further enhances the fine-tuning performance of the existing AT methods. In iDAT, smaller model, such as ViT-S, assume the role of teacher, while larger model, like ViT-B, function as student. This approach is contrary to the traditional knowledge distillation, which usually involves larger models imparting knowledge to smaller or equally-sized models, hence the term inverse distillation. Utilizing the broader adapter weight distribution of the smaller model provides the larger model with diverse perspectives of knowledge. This aids the latter in comprehensively acquiring downstream knowledge and better adapting to downstream tasks. Moreover, compared to traditional knowledge distillation methods, our inverse distillation approach achieves superior results with fewer additional trainable parameters, making it more suitable for AT methods.

Extensive experiments are conducted on the VTAB-1K benchmark, encompassing 19 distinct image classification tasks, to validate the efficacy of our proposed framework. The results demonstrate that the implementation of iDAT can further enhance the performance of AT methods on downstream tasks. Specifically, employing the Sequential Adapter \cite{nlpadapter} as AT method, a performance gain of 2.66\% is achieved within our iDAT framework. Remarkably, this surpasses recent state-of-the-art methods, such as Res-Tuning \cite{Res-Tuning}, by 0.12\%, while only utilizing 36\% of the latter's parameter count. Hence, our iDAT framework proves to be a simple yet effective approach for augmenting existing AT methods, enabling them to rival state-of-the-art performances without bells and whistles.

The contributions of this paper are summarized as follows: \textbf{1)} By visualizing the weight distribution within the linear layers of the Adapter module, we conduct analysis of the knowledge discrepancy between models of varying sizes after AT, offering novel insights into knowledge transfer. \textbf{2)} Leveraging the knowledge discrepancy represented by the weight distribution differences, we naturally develop a knowledge distillation framework suitable for AT, named iDAT. This framework innovatively utilizes small models to provide distinct perspectives of knowledge to enhance the downstream fine-tuning performance of larger models. \textbf{3)} We conduct extensive experiments on 19 datasets, demonstrating the effectiveness of iDAT across different Adapter-Tuning methods. The results indicate that, based on iDAT, even the most simplistic Adapter modules can match the performance of state-of-the-art approaches in a straightforward manner.

\section{Related Works}
\label{sec:relatedworks}

\subsection{Adapter-Tuning}

As a parameter-efficient fine-tuning method, Adapter-Tuning (AT) freezes the pre-trained model and introduces learnable adapter modules to acquire downstream knowledge, which is popular due to its efficiency and plug-and-play characteristics. Figure \ref{fig:adapter} illustrates three classic AT methods, categorized as Sequential Adapter, Parallel Adapter, and Parallel Shared Adapter, contingent upon the varying insertion positions of the adapter module. Additionally, there are some more carefully designed methods such as Mona \cite{Mona} and MoSA \cite{MoSA}. Mona enhances its ability to process visual signals by introducing multiple visually friendly filters. MoSA employs mixture-of-experts for sparse training to fully leverage the potential of each parameter in the adapter modules.

\subsection{Knowledge distillation}

Knowledge distillation is a popular technique that involves transferring the knowledge from large-scale models to smaller-scale models in a distillation process, widely applied across various tasks \cite{kdsurvey}. Online distillation, a variant of knowledge distillation, facilitates the transfer of knowledge from teacher models to student models in a more user-friendly manner by training them simultaneously \cite{sfokd, peftokd}. However, most existing knowledge distillation methods rely on the premise that larger models inherently contain more knowledge than smaller counterparts. In this paper, via observations from preliminary experiments, we explore inverse distillation and propose iDAT, a distillation framework better suited for AT methods.

\section{Preliminary}

\begin{figure}[!t]
    \centering
    \includegraphics [width=0.48\textwidth] {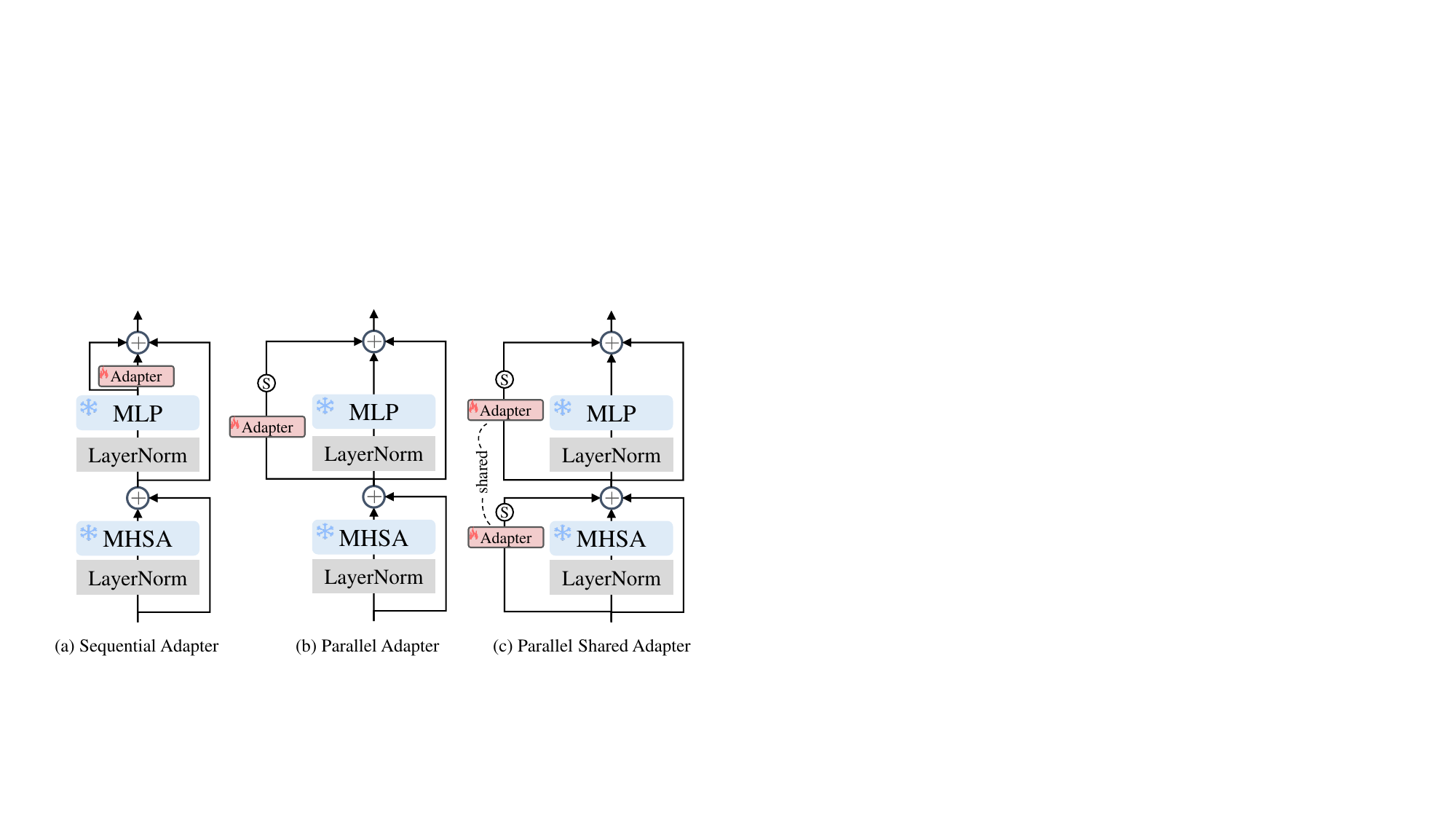}
    \caption{Three classical adapter structures.}
    \label{fig:adapter}
\end{figure}

The classic work on Sequential Adapter is \cite{nlpadapter}, as shown in Figure \ref{fig:adapter}(a). The authors sequentially introduce adapter modules with residual connections after the MLP layer in TransFormer blocks to adjust the output, adapting it to downstream tasks, which can be represented in Equation \ref{eq:seqadapter}.

\begin{equation}
    x^{\prime} = W_{up}(\sigma(W_{down}(x))) + x
    \label{eq:seqadapter}
\end{equation}

where $W_{down} \in \mathbf{R}^{d \times d^{\prime}}, W_{up} \in \mathbf{R}^{d^{\prime} \times d}$ are both linear layers, with $d^{\prime} \ll d$. $\sigma$ represents the activation function, and in this paper, GELU is used as the activation function.

A representative study of Parallel Adapter is AdaptFormer \cite{adaptformer}, as shown in Figure \ref{fig:adapter}(b). It places adapter modules in parallel alongside the MLP and introduces a scaling factor to control the impact of adapter outputs on the backbone, which can be represented as Equation \ref{eq:adaptformer}. Subsequently, as shown in Figure \ref{fig:adapter}(c), Parallel Shared Adapter \cite{GIST} incorporates parameter sharing based on the Parallel Adapter, introducing the same adapter module alongside both MHSA and MLP pathways, enhancing the efficiency of parameter utilization.

\begin{equation}
    x^{\prime} = s * W_{up}(\sigma(W_{down}(x)))
    \label{eq:adaptformer}
\end{equation}
where $s$ represents the scaling factor, which is set to 0.1 by default in \cite{adaptformer} and this paper.

\section{Methods}
\label{sec:methods}

In Section 4.1, we first analyze the weight distribution of adapter modules in ViT models of different scales, uncovering weight disparities and the knowledge differences they represent. Subsequently, based on the differences, we explore the possibility of applying online distillation in the AT methods. In Section 4.2, we propose the inverse Distillation Adapter-Tuning, aiming to unleash the potential of existing AT methods on downstream fine-tuning.

\begin{table}[!t]
    \centering
    \scriptsize
    \begin{tabular}{c|ccc|c}
    \hline
          Model & Size (M) & Tr. Params. (M) &Mem. (GB) & Acc. (\%) \\
         \hline
         ViT-S & 22.0 & 0.05&4.8 &  80.42\\
         ViT-B & 86.6 & 0.09&6.8 & 81.42 \\
          ViT-L & 307.0 & 0.25&13.2 & 81.96 \\
         \hline
    \end{tabular}
    \caption{Mean Accuracy on CIFAR-100, DTD and Flowers102 datasets. Parallel Adapter is utilized as the AT method. \textit{Size} stands for backbone parameters, \textit{Tr. Params.} represents trainable parameters introduced by adapter modules, and \textit{Mem.} represents the GPU memory consumption.}
    \label{tab:three-vit-paralleldadapter}
\end{table}

\subsection{Empirical experiments}

In this section, the Parallel Adapter \cite{adaptformer} is employed as the AT method to fine-tune three scales of ViT models on the CIFAR-100, DTD and Flowers102 dataset, as shown in Table \ref{tab:three-vit-paralleldadapter}. The gap in fine-tuning performance suggests differences in the downstream knowledge acquired by the adapter modules of different models. To further explore this discrepancy, a visualization of the weight distribution in the linear layers \(W_{down}\) and \(W_{up}\) of the fine-tuned adapter module is conducted\footnote{In this section, we only display the weight distribution of the first layer's adapter module in the ViT model. More visualizations could be found in the supplementary materials.}, as depicted in Figure \ref{fig:motivation}. It is apparent that in the larger-scale models (ViT-B and ViT-L), the weights of the adapter module are more concentrated and exhibit a narrow distribution, indicating pronounced sparsity. Conversely, in the smaller-scale model (ViT-S), the adapter module weights are more dispersed and demonstrate a flatter distribution, indicative of less sparsity. This variation in weight distribution further corroborates the notion that the knowledge acquired by the adapter modules varies across different models. Specifically, in the ViT-B and ViT-L models, the downstream knowledge learned by the adapter modules is more similar, whereas in the ViT-S model, it significantly diverges from the former. This observation raises a pertinent question: \textit{Can this natural disparity in knowledge contribute diverse perspectives to the learning process of the models, thereby enriching the acquisition of downstream knowledge?}

\begin{figure}[!t]
    \centering
    \includegraphics [width=0.48\textwidth] {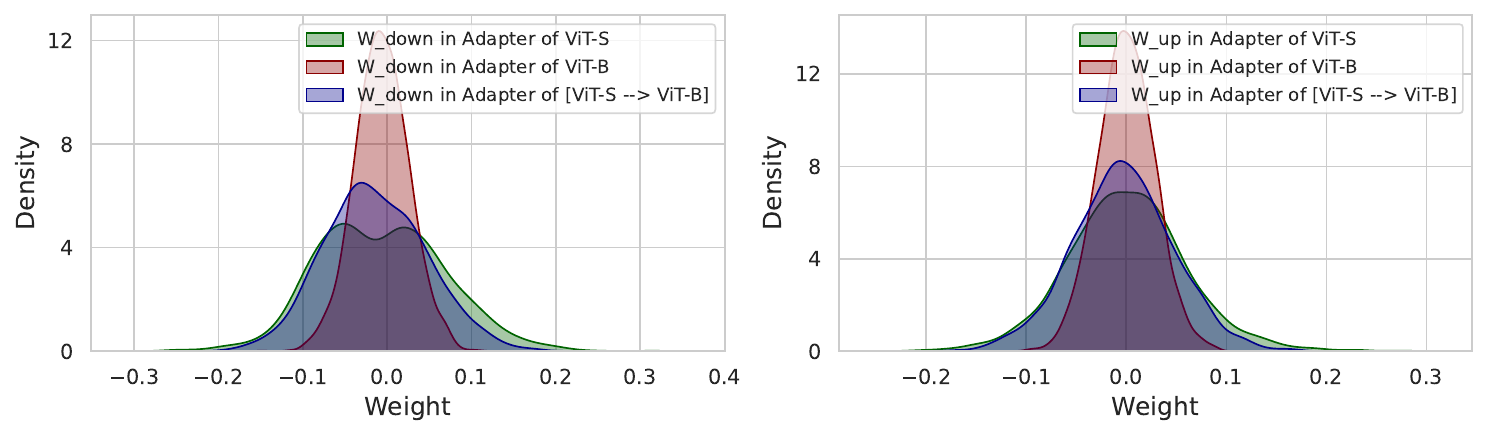}
    \caption{Weight distribution visualization of adapter modules on  CIFAR-100 dataset.}
    \label{fig:w_after}
\end{figure}

\begin{table}[!t]
    \centering
    \scriptsize
    \begin{tabular}{cc|ccc}
    \hline
          Teacher & Student & Tr. Params. (M) &Mem. (GB) & Acc. (\%) \\
         \hline
         ViT-S & ViT-B & \textbf{0.14} &\textbf{8.3} & \textbf{83.37} \\
         ViT-B & ViT-B & 0.18 &10.2 & 82.11 \\
          ViT-L & ViT-B & 0.34 &16.7 & 83.03 \\
         \hline
    \end{tabular}
    \caption{Mean Accuracy on CIFAR-100, DTD and Flowers102 datasets. Parallel Adapter is utilized as the AT method.}
    \label{tab:empiricalexp}
\end{table}

To address this question, empirical experiments are conducted. Utilizing Parallel Adapter as the AT method, and employing ViT-B as the student model, online distillation is performed with ViT-S, ViT-B, and ViT-L serving as teacher models. Kullback-Leibler divergence loss is utilized as the distillation loss, and the results are presented in Table \ref{tab:empiricalexp}. Surprisingly and intriguingly, the most effective result is achieved when using the weakest model (ViT-S) as the teacher for online distillation. This finding addresses the aforementioned question. In ViT-S, the weight distribution of the adapter module is more dispersed, whereas, in ViT-B and ViT-L, it is more concentrated. This disparity allows ViT-S, when acting as a teacher, to impart a different perspective of knowledge to the student model ViT-B, thereby facilitating a more comprehensive acquisition of downstream knowledge. Subsequently, we visualize the weight distribution of the adapter module when ViT-S served as the teacher and ViT-B as the student, as shown in Figure \ref{fig:w_after}. It is observed that the concentration of the weight distribution in the adapter module of ViT-B became less pronounced, indicating that ViT-S indeed injects different knowledge into ViT-B. Moreover, employing a smaller model as the teacher also significantly reduces fine-tuning overheads, as shown in Table \ref{tab:three-vit-paralleldadapter} and \ref{tab:empiricalexp}. This approach decreases the trainable parameters and GPU memory usage by about 50\% compared to traditional knowledge distillation (ViT-L $\rightarrow$ ViT-B), and only resulted in 0.05M additional trainable parameters and a 20\% increase in memory usage compared to solely fine-tuning ViT-B.

\subsection{inverse Distillation Adapter-Tuning (iDAT)}

\begin{figure}[!t]
    \centering
    \includegraphics [width=0.48\textwidth] {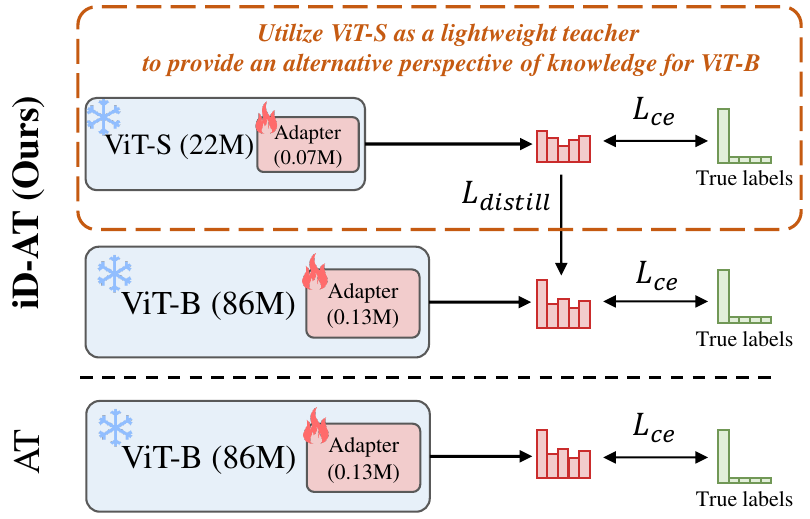}
    \caption{inverse Distillation Adapter-Tuning.}    
    \label{fig:main}
\end{figure}

Inspired by the above observations, we propose inverse Distillation Adapter-Tuning, as shown in Figure \ref{fig:main}. Unlike traditional knowledge distillation, which transfers knowledge from larger to smaller models, we utilize a smaller model (\textit{e.g.}, ViT-S) as the teacher to provide a different perspective of knowledge to enhance the performance of the larger model (\textit{e.g.}, ViT-B). Depending on the Adapter-Tuning (AT) method and loss function, we offer six variants: iDAT-S-mse, iDAT-P-mse, iDAT-PS-mse, iDAT-S-kl, iDAT-P-kl, and iDAT-PS-kl. Where S denotes the use of adapter modules is Sequential Adapter \cite{nlpadapter}, as shown in Figure \ref{fig:adapter} (a). P indicates parallel placement, as in Figure \ref{fig:adapter} (b) \cite{adaptformer}. PS signifies a parameter-sharing form of parallel placement, as in Figure \ref{fig:adapter} (c) \cite{GIST}. Besides, mse and kl represent the distillation loss, as illustrated in Equation \ref{eq:mseloss} and \ref{eq:klloss}.

\begin{equation}
\mathcal{L}_{mse}(\hat{y_s},\hat{y_t})=(\hat{y_s}-\hat{y_t})^2 
\label{eq:mseloss}
\end{equation}
\begin{equation}
\mathcal{L}_{kl}(\hat{y_s},\hat{y_t}; \mathcal{T})=\textup{softmax}(\hat{y_s})\textup{log}\frac{\textup{softmax}(\hat{y_s}/\mathcal{T})}{\textup{softmax}(\hat{y_t}/\mathcal{T})}
\label{eq:klloss}
\end{equation} where, $\hat{y_s}, \hat{y_t}$ respectively represent the outputs of the student and teacher. $\mathcal{T}$ is a temperature coefficient to soften the outputs \cite{kdsurvey}.

Notably, only the student model is used during inference for saving parameters and deployment costs, avoiding any additional burden. Depending on the loss function used for knowledge transfer during training, our total loss function is defined as in Equation \ref{eq:allloss}.

\begin{equation}
\begin{aligned}
    \mathcal{L}_{basic}=\mathcal{L}_{ce}(\hat{y_s},y)+\mathcal{L}_{ce}(\hat{y_t},y) \\
    \mathcal{L}_{all-mse}=\mathcal{L}_{basic}+\lambda_{mse}\mathcal{L}_{mse}(\hat{y_s},\hat{y_t}) \\
    \mathcal{L}_{all-kl}=\mathcal{L}_{basic}+\lambda_{kl}\mathcal{L}_{kl}(\hat{y_s},\hat{y_t}; \mathcal{T})
    \label{eq:allloss}
\end{aligned}
\end{equation} where, $y$ represents the true labels, $\mathcal{L}_{basic}$ denotes the update of both the student and teacher models using cross-entropy loss, $\lambda_{mse},\lambda_{kl}$ represent the weight coefficients, respectively. In this paper, $\lambda_{mse}$ and $\lambda_{kl}$ are set to 1.0, and the temperature coefficient $\mathcal{T}$ is set to 5.0 by default.

\begin{table*}[!t]
\centering
\scriptsize
\setlength{\tabcolsep}{1.3pt}
\begin{tabular}{cc|ccccccc|cccc|cccccccc|cc}
\hline
\multicolumn{2}{l|}{}  & \multicolumn{7}{c|}{Natural}                                      & \multicolumn{4}{c|}{Specialized}                  & \multicolumn{8}{c|}{Structured}                                                                                     & \multicolumn{1}{l}{} & \multicolumn{1}{l}{} \\ \hline
Method & Comments  & \rotatebox{90}{CIFAR-100} & \rotatebox{90}{Caltech101} & \rotatebox{90}{DTD}  & \rotatebox{90}{Flowers102} & \rotatebox{90}{Pets} & \rotatebox{90}{SVHN} & \rotatebox{90}{Sun397} & \rotatebox{90}{Patch Camelyon} & \rotatebox{90}{EuroSAT} & \rotatebox{90}{Resisc45} & \rotatebox{90}{Retinopathy} & \rotatebox{90}{Clevr/count} & \rotatebox{90}{Clevr/distance} & \rotatebox{90}{DMLab} & \rotatebox{90}{KITTI/distance} & \rotatebox{90}{dSprites/loc} & \rotatebox{90}{dSprites/ori} & \rotatebox{90}{SmallNORB/azi} & \rotatebox{90}{SmallNORB/ele} & \rotatebox{90}{Mean Acc. (\%)}        & \rotatebox{90}{Tr. Params. (M)}          \\ \hline
FPT          & -          & 68.9      & 87.7       & 64.3 & 97.2       & 86.9 & 87.4 & 38.8   & 79.7           & 95.7    & 84.2     & 73.9        & 56.3        & 58.6           & 41.7  & 65.5           & 57.5         & 46.7         & 25.7          & 29.1          & 65.57          & 85.84                \\
Linear Probing     & -   & 63.4      & 85.0         & 63.2 & 97.0         & 86.3 & 36.6 & 51.0     & 78.5           & 87.5    & 68.6     & 74.0          & 34.3        & 30.6           & 33.2  & 55.4           & 12.5         & 20.0           & 9.6           & 19.2          & 52.94         & 0.04                 \\ \hline
VPT-Deep \cite{VPT} & ECCV22 & 78.8 & 90.8 &65.8 & 98.0&88.3 &78.1 &49.6 &81.8 &96.1 &83.4 &68.4 &68.5 &60.0 &46.5 &72.8 &73.6 &47.9 &32.9 &37.8 &69.43 &0.60  \\ 
LoRA \cite{nlplora} & ICLR21&65.3&87.9&69.4&98.7&90.7&82.4&53.4&82.8&94.8&82.5&75.0&77.6&64.7&45.8&79.0&73.3&44.7&26.3&38.2&70.13&0.29 \\
SSF        \cite{ssf}   & NeurIPS22     & 69.0        & 92.6       & 75.1 & 99.4       & 91.8 & 90.2 & 52.9   & \textbf{87.4}           & 95.9    & 87.4     & 75.5        & 75.9        & 62.3           & 53.3  & 80.6           & 77.3         & 54.9        & 29.5          & 37.9          & 73.10      & 0.24                 \\
ARC \cite{ARC} & NeurIPS23 &72.2&90.1&72.7&99.0&91.0&\textbf{91.9}&54.4&84.9&95.7&86.7&75.8&80.7&67.1&48.7&81.6&79.2&51.0&31.4&39.9&73.37&0.13 \\
MoSA \cite{MoSA} & Arxiv23 &\textbf{79.7} &91.5 &66.2 &98.8 &89.7 &79.0 &53.4 &83.4 &95.6 &82.0 &75.1 &71.5 &58.1 &40.7 &70.2 &57.8 &43.6 &26.5 &34.0 &68.25 &0.30 \\ 
ReAdapter \cite{readapter} &Arxiv23 &72.4 &91.6 &71.0 &99.2 &91.4 &90.7 &55.1 &85.3 &95.9 &84.6 &75.9 &82.3 &\textbf{68.0} &50.4 &79.9 &80.4 &49.2 &\textbf{38.6}&41.0 &73.84 &0.22 \\
Res-Tuning \cite{Res-Tuning} & NeurIPS23    &75.2 &92.7 &71.9 &99.3 &91.9 &86.7 &\textbf{58.5} &86.7 &95.6 &85.0 &74.6 &80.2 &63.6 &50.6 &80.2 &\textbf{85.4} &\textbf{55.7} &31.9 &42.0 &74.09 & 0.55 \\ \hline

Sequential Adapter* \cite{nlpadapter} &ICML19 &70.1 	&93.5 	&74.9 	&99.5 	&91.7 	&87.2 	&51.4 	&86.9 	&96.6 	&87.8 	&76.9 	&84.3 	&34.5 	&53.8 	&80.2 	&72.8 	&54.8 	&22.4 	&40.2  & 71.55 &0.13\\
iDAT-S-kl &Ours &75.3 	&93.7 	&76.7 	&\textbf{99.6} 	&92.3 	&88.6 	&55.1 	&86.8 	&\textbf{96.9} 	&\textbf{88.7} 	&\textbf{77.0} 	&85.1 	&62.6 	&53.7 	&79.4 	&77.5 	&54.7 	&27.3 	&38.9   &\textbf{74.21}  &0.20  \\ 
iDAT-S-mse &Ours  &76.0 	&\textbf{94.7} 	&77.2 	&\textbf{99.6} 	&92.3 	&90.0 	&55.5 	&86.7 	&96.6 	&88.4 	&\textbf{77.0} 	&\textbf{86.4} 	&29.0 	&52.8 	&\textbf{82.3} 	&80.3 	&54.5 	&29.3 	&\textbf{43.5}  & 73.27 &0.20  \\ 
\hline
Parallel Adapter* \cite{adaptformer} &NeurIPS22 &70.4 	&92.2 	&74.5 	&99.4 	&91.3 	&79.1 	&51.5 	&83.2 	&96.4 	&87.7 	&76.2 	&83.7 	&60.3 	&53.5 	&74.6 	&56.6 	&54.4 	&28.3 	&42.3  &71.35 &0.13 \\
iDAT-P-kl &Ours &73.7 	&92.3 	&76.8 	&\textbf{99.6} 	&92.1 	&83.5 	&55.0 	&84.1 	&96.5 	&88.2 	&76.7 	&82.1 	&60.6 	&53.4 	&76.3 	&67.1 	&53.9 	&29.3 	&40.7 &72.73   &0.20  \\
iDAT-P-mse &Ours &74.6 	&94.0 	&77.0 	&\textbf{99.6} 	&\textbf{92.5} 	&85.9 	&55.2 	&85.1 	&96.6 	&88.0 	&\textbf{77.0} 	&80.3 	&62.1 	&53.2 	&73.9 	&76.0 	&55.1 	&30.1 	&41.6  & 73.57&0.20 \\
\hline
Parallel Shared Adapter \cite{GIST} &Arxiv23  & 70.2      & 92.6       & 74.6 & 99.4       & 91.2 & 80.4 & 51.4   & 84.1           & 96.3    & 88.0       & 75.6        & 84.2        & 59.6           & 53.2  & 76.3           & 60.7         & 51.9         & 27.8          & 40.2          & 71.46    & 0.13 \\
iDAT-PS-kl &Ours  &73.9 	&92.8 	&76.9 	&\textbf{99.6} 	&92.3 	&84.0 	&54.9 	&85.1 	&\textbf{96.9} 	&88.6 	&76.2 	&83.7 	&60.4 	&53.4 	&77.4 	&70.7 	&53.2 	&27.2 	&40.2 &73.02  &0.20  \\ 
iDAT-PS-mse &Ours &74.9 	&94.1 	&\textbf{77.5} 	&\textbf{99.6}	&92.4 	&87.1 	&55.2 	&85.1 	&96.6 	&88.6 	&76.9 	&83.1 	&62.4 	&\textbf{54.9} 	&73.2 	&79.0 	&54.7 	&29.5 	&41.3  & 74.00 & 0.20  \\ 
\hline

\end{tabular}
\caption{\textbf{The comparative results on the VTAB-1K benchmark} with the ViT-B/16 model pre-trained on ImageNet-21K. Top-1 Mean Accuracy over 19 datasets is reported. \textit{Tr. Params.} stands for trainable parameters. * Due to the absence of results in \cite{nlpadapter, adaptformer}, we re-implement them in our settings. The best results are in \textbf{bold}.}
\label{tab:compare}
\end{table*}

\begin{table}[t]
\begin{minipage}[c]{0.48\linewidth}
\centering
\scriptsize
    \begin{tabular}{cc}
    \hline
          Loss & Acc. (\%) \\
         \hline
         - &70.37 \\
         $\mathcal{L}_{mae}$  & 74.89 \\
         $\mathcal{L}_{cos}$  & 73.10 \\
         \hline
    \end{tabular}
\caption{Ablations on different loss functions.}
\label{tab:kdloss}
\end{minipage}
\hfill
\begin{minipage}[c]{0.48\linewidth}
    \centering
    \scriptsize
    \begin{tabular}{cc}
    \hline
          Distill. framework & Acc. (\%) \\
    \hline
            - & 70.37 \\
          Ours & 73.73\\
          feature-based distill. & 66.21 \\
    \hline
    \end{tabular}
    \caption{Ablations on different distillation paradigms.}
    \label{tab:Various distillation paradigms}
\end{minipage}
\end{table}

\begin{table}[]
    \centering
    \scriptsize
    \setlength{\tabcolsep}{2pt}
    \begin{tabular}{c|cc|cc|cc|cc}
    \hline
    &\multicolumn{2}{c}{h=8} &\multicolumn{2}{c}{h=16}&\multicolumn{2}{c}{h=32}&\multicolumn{2}{c}{h=64} \\ \hline
   Methods & \rotatebox{90}{Acc. (\%)} &\rotatebox{90}{Tr. Params. (M)} & \rotatebox{90}{Acc. (\%)} &\rotatebox{90}{Tr. Params. (M)} & \rotatebox{90}{Acc. (\%)} &\rotatebox{90}{Tr. Params. (M)} & \rotatebox{90}{Acc. (\%)} &\rotatebox{90}{Tr. Params. (M)} \\ \hline
   ViT-B & 70.80&0.23 &70.08 &0.38 &71.84 & 0.68&70.87 &1.27 \\
   ViT-L & 73.15&0.52 &73.86 &0.91 &73.48 &1.70 &73.40 &3.27\\
   iDAT-P-kl & 74.20&0.35 &74.03 &0.57 &75.14 &1.01 &75.20 &1.90\\
   iDAT-P-mse &75.05 &0.35 &75.76 &0.57 &76.47 &1.01 &77.44 &1.90\\
    \hline
    \end{tabular}
    \caption{Ablations on hidden dimension of adapter modules.}
    \label{tab:hiddensize}
\end{table}

\section{Experiments}
\label{sec:experiments}

\subsection{Experimental settings}
\label{sec:experimentalsettings}
We validate six variants of our iDAT framework on the Visual Task Adaptation Benchmark (VTAB-1K) \cite{vtab1k}. VTAB-1K comprises 19 image classification tasks, categorized into three groups: natural, specialized, and structured, covering different domains. Each sub-task contains only 1,000 training samples and an average of 20,000 test samples, with an average of 50 categories, which makesVTAB-1K highly challenging. Following previous work \cite{ssf}, we report the Top-1 accuracy for all 19 tasks, as well as the average accuracy across the entire benchmark. For the training setup, we follow all the settings as outlined in \cite{ssf}. For our iDAT framework, the ViT-B/16, pre-trained on ImageNet-21K, is employed as the backbone network, while the ViT-S/16 served as the teacher model. The hidden dimension of the adapter module is set to 4, and the scaling factor is 0.1 unless specified otherwise. Experiments are conducted on NVIDIA A100 GPUs.

\subsection{Main results}

As illustrated in Table \ref{tab:compare}, extensive comparative experiments have been conducted on the VTAB-1K dataset. Remarkably, with an addition of merely 0.07M parameters, our iDAT framework is capable of enhancing the performance of three plain baselines by an average of 2.01\%, which underscores the universality of our framework. Furthermore, compared to certain meticulously designed methods (\textit{e.g.} MoSA, ReAdapter, and Res-Tuning), our iDAT-S-kl achieves superior fine-tuning performance with fewer trainable parameters. This not only demonstrates the simplicity and effectiveness of our framework, but also its ability to rival state-of-the-art performances without bells and whistles.

\subsection{Ablation studies}

In this section, the ablation studies are conducted on CIFAR-100 dataset. Unless specifically indicated otherwise, we employ the ViT-B/16 and ViT-S/16 pre-trained on ImageNet-21K as the student and teacher model, respectively. The Parallel Adapter \cite{adaptformer} is utilized as the AT method, and the Kullback-Leibler divergence loss is used as the distillation loss. For the baseline, fine-tuning is conducted solely with the Parallel Adapter without additional knowledge integration.

\textbf{Hyperparameters in iDAT.} The ablation studies on $\lambda_{kl}$, $\lambda_{mse}$ and the temperature $\mathcal{T}$ of $\mathcal{L}_{kl}$ are in the appendix.

\textbf{More distillation loss functions.} As shown in Table \ref{tab:kdloss}, we explore the loss functions employed for distillation. Beyond the Kullback-Leibler divergence loss and Mean Square Error loss, we have also experimented with Cosine Similarity loss and Mean Absolute Error loss. The results underscore the scalability of the iDAT framework, demonstrating that various distillation losses can lead to performance improvements.

\textbf{Various distillation paradigms.} We delve into more intricate distillation paradigms, such as the feature-based distillation paradigm \cite{Fitnets}, as demonstrated in Table \ref{tab:Various distillation paradigms}. The results indicate that the more complicated distillation paradigm can even lead to a decline in performance and are not compatible with Adapter-Tuning. Consequently, our iDAT framework employs the simplest and most direct logit-based distillation paradigm, which enhances the fine-tuning performance of existing AT methods in an elegantly simple manner.

\textbf{Larger adapter modules.} Table \ref{tab:hiddensize} examines the effects of dimension scaling in the adapter module. The intermediate layer dimensions are varied from 4 to 8, 16, 32, and 64. Observations reveal an enhancement in the iDAT framework's fine-tuning performance with increasing intermediate layer dimensions, underscoring the framework's capability to achieve substantial gains in diverse parameter counts scenarios. Notably, with the hidden layer dimension set to 64, the iDAT framework significantly elevates ViT-B's fine-tuning performance from 70.87\% to 77.44\%, utilizing merely an additional 0.63M trainable parameters.

\begin{table}[]
    \centering
    \scriptsize
    \begin{tabular}{cc|cc}
    \hline
          Teacher & Student & Tr. Params. (M) & Acc. (\%) \\
    \hline
          - & ViT-S &0.08&68.26 \\
          - & ViT-L &0.32&72.45 \\
        ViT-L & ViT-L &0.64&74.89 \\
          \hline
          ViT-S & ViT-L &0.40&75.74 \\
    \hline
    \end{tabular}
    \caption{Ablations on other teacher-student pairings.}
    \label{tab:More teacher-student pairings}
\end{table}

\textbf{More teacher-student pairings.} Adhering to a small-to-large distillation concept in iDAT, we have validated another small teacher-large student pairing (ViT-S $\rightarrow$ ViT-L), as shown in Table \ref{tab:More teacher-student pairings}. The results indicate that our framework is not limited to distillation from ViT-S to ViT-B.

\begin{table}[]
    \centering
    \scriptsize
    \begin{tabular}{cc|cc}
    \hline
          Teacher & Student & Tr. Params. (M) & Acc. (\%) \\
    \hline
          - & Swin-S &0.16&68.49 \\
          - & Swin-B &0.21&69.86 \\
          \hline
          Swin-S & Swin-B &0.37&71.72 \\
    \hline
    \end{tabular}
    \caption{Ablations on Swin family models.}
    \label{tab:swin}
\end{table}

\textbf{Various model families.} We employ the Swin \cite{swinTRM} models as a substitute for ViT models as the backbone, as shown in Table \ref{tab:swin}. The results indicate that our iDAT framework is also applicable to Swin models, utilizing Swin-S as a lightweight teacher to provide diversified knowledge for the larger Swin-B, thereby enhancing fine-tuning performance. This further underscores the universality of iDAT framework.

\subsection{Visualization}

\begin{figure}[!t]
    \centering
    \includegraphics [width=0.48\textwidth] {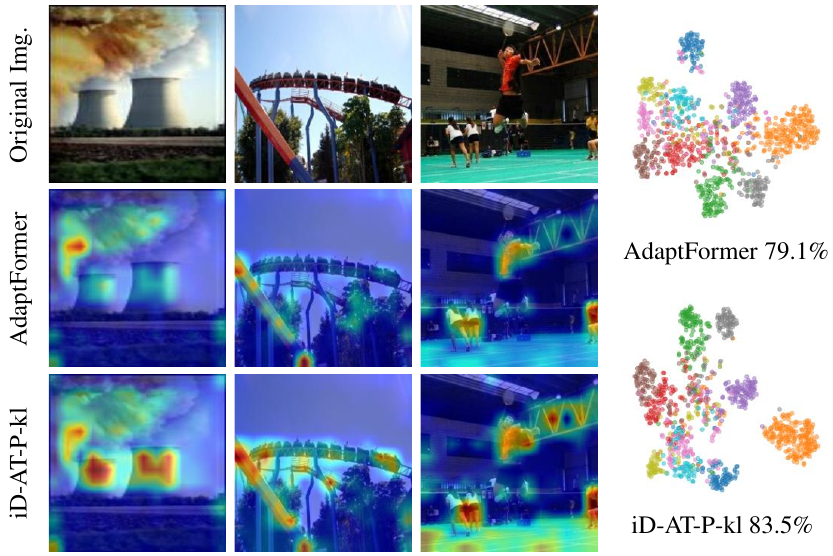}
    \caption{\textbf{Left}: Attention map visualization on Sun397 dataset. \textbf{Right}: t-SNE visualization on SVHN dataset.}    
    \label{fig:vis}
\end{figure}

In this section, attention maps and t-SNE visualizations are conducted using the [CLS] token extracted from the final layer of the Transformer block, preceding the linear classification head, as depicted in Figure \ref{fig:vis}. This further elucidate the efficacy of our framework.

\section{Conclusions}
\label{sec:conclusions}

In this paper, the feasibility of incorporating a knowledge distillation framework within the Adapter-Tuning (AT) methodology is investigated, marking a pioneering effort in this domain. Intriguingly, empirical experiments reveal that the employment of a relatively weak and small model as the teacher paradoxically yields superior fine-tuning outcomes, with the added benefit of requiring only a limited number of additional learnable parameters. Stemming from these findings, our inverse Distillation Adapter-Tuning framework is introduced, aimed at improving the fine-tuning performance of existing AT methods. Our iDAT framework employs a smaller model as a lightweight teacher, providing knowledge from diverse perspectives to enhance the performance of the larger model. Comprehensive experimental validation underscores the effectiveness, universality, and scalability of iDAT. Future endeavors will concentrate on developing a knowledge distillation framework that is more adept at facilitating parameter-efficient fine-tuning techniques.

\bibliographystyle{IEEEbib}
\bibliography{mycite}

\clearpage
\appendix

\section{More ablation studies}

In this section, the ablation studies are conducted on CIFAR-100 dataset. Unless specifically indicated otherwise, we employ the ViT-B/16 and ViT-S/16 pre-trained on ImageNet-21K as the student and teacher model, respectively. The Parallel Adapter \cite{adaptformer}, is utilized as the AT method, and the Kullback-Leibler divergence loss is used as the distillation loss. For the baseline, fine-tuning is conducted solely with the Parallel Adapter as the AT approach, devoid of additional knowledge integration.

\begin{table}[h]
\centering
    \setlength{\tabcolsep}{2pt}
    \begin{tabular}{cc}
    \hline
          $\mathcal{T}$ & Top1-Acc. (\%) \\
         \hline
         1   & 66.66 \\
         5   & 73.62 \\
         10  & 72.65 \\
         20  & 71.15 \\
         \hline
    \end{tabular}
\caption{Ablations on temperature $\mathcal{T}$ of $L_{kl}$.}
\label{tab:temp}
\end{table}

\begin{table}[h]
\centering
    \setlength{\tabcolsep}{2pt}
    \begin{tabular}{cc}
    \hline
          $\lambda_{kl}$ & Top1-Acc. (\%) \\
         \hline
         0.1   & 72.02 \\
         0.2   & 72.80 \\
         0.5  & 73.62 \\
         1.0  & 73.73 \\
         \hline
    \end{tabular}
\caption{Ablations on $\lambda_{kl}$.}
\label{tab:lambdakl}
\end{table}

\begin{table}[h]
\centering
    \setlength{\tabcolsep}{2pt}
    \begin{tabular}{cc}
    \hline
          $\lambda_{mse}$ & Top1-Acc. (\%) \\
         \hline
         0.1   & 74.24 \\
         0.2   & 74.34 \\
         0.5  & 74.57 \\
         1.0  & 74.61 \\
         \hline
    \end{tabular}
\caption{Ablations on $\lambda_{mse}$.}
\label{tab:lambdamse}
\end{table}

\textbf{Hyperparameters in iDAT.} The ablation studies focusing on hyperparameters within the iDAT-P-kl framework, delineated in Tables \ref{tab:temp} and \ref{tab:lambdakl}, reveal the most effective settings to be a temperature coefficient of 5 and a weight coefficient of 0.5. Additionally, as evidenced in Table \ref{tab:lambdamse}, the iDAT-P-mse framework's weight coefficients are similarly examined. These results indicate a comparatively lower hyperparameter sensitivity for the iDAT-P-mse framework than for iDAT-P-kl, positioning mse loss as a more robust option for distillation.

\section{More Visualization}

\begin{figure*}
    \centering
    \includegraphics[width=0.98\textwidth]{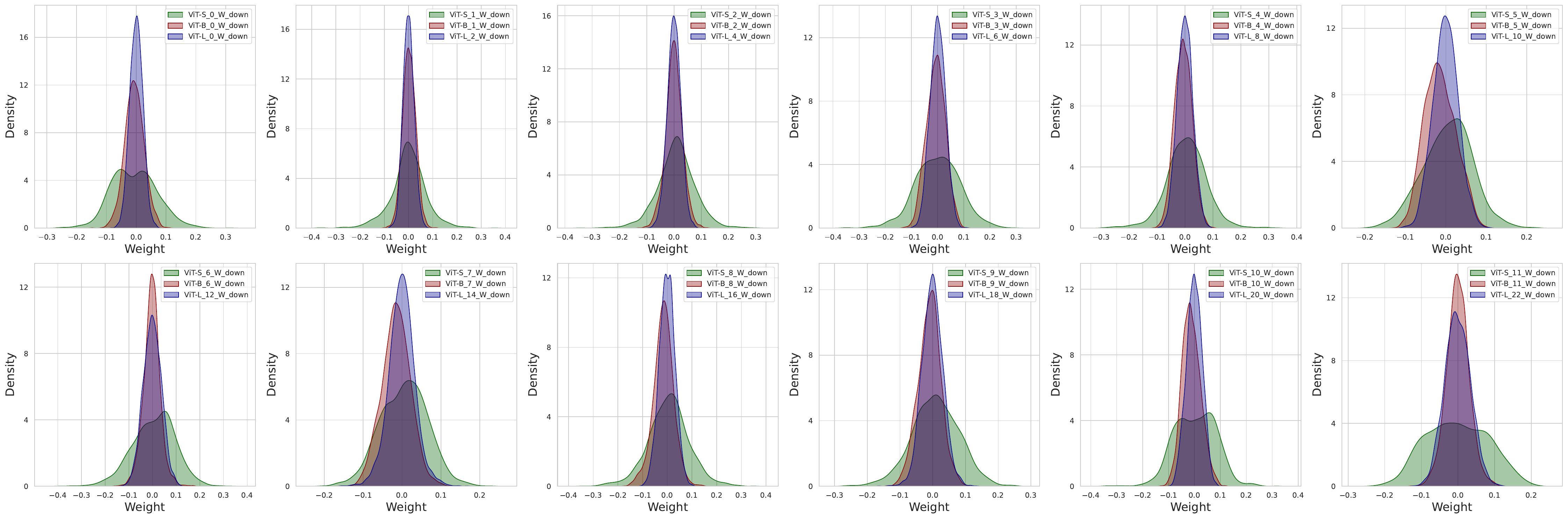}
    \caption{Visualization of the weight distribution of $W_{down}$ in the adapter module on CIFAR-100 dataset.}
    \label{fig_all_wdown_before}
\end{figure*}

\begin{figure*}
    \centering
    \includegraphics[width=0.98\textwidth]{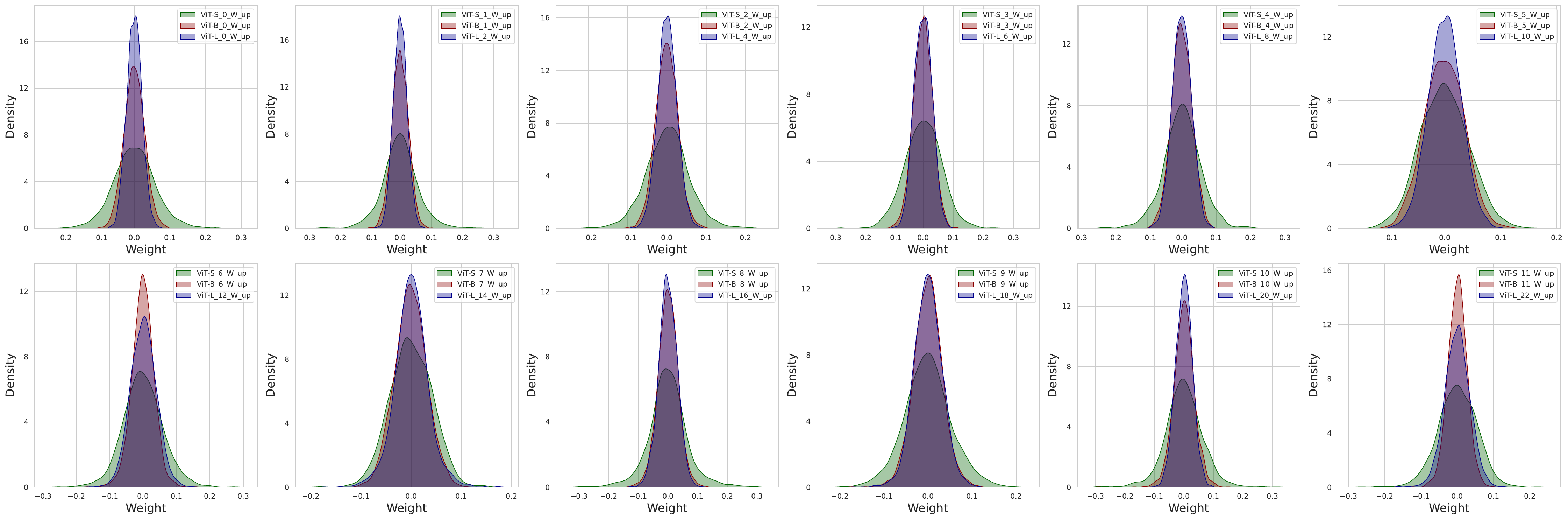}
    \caption{Visualization of the weight distribution of $W_{up}$ in the adapter module on CIFAR-100 dataset.}
    \label{fig_all_wup_before}
\end{figure*}

\begin{figure*}
    \centering
    \includegraphics[width=0.98\textwidth]{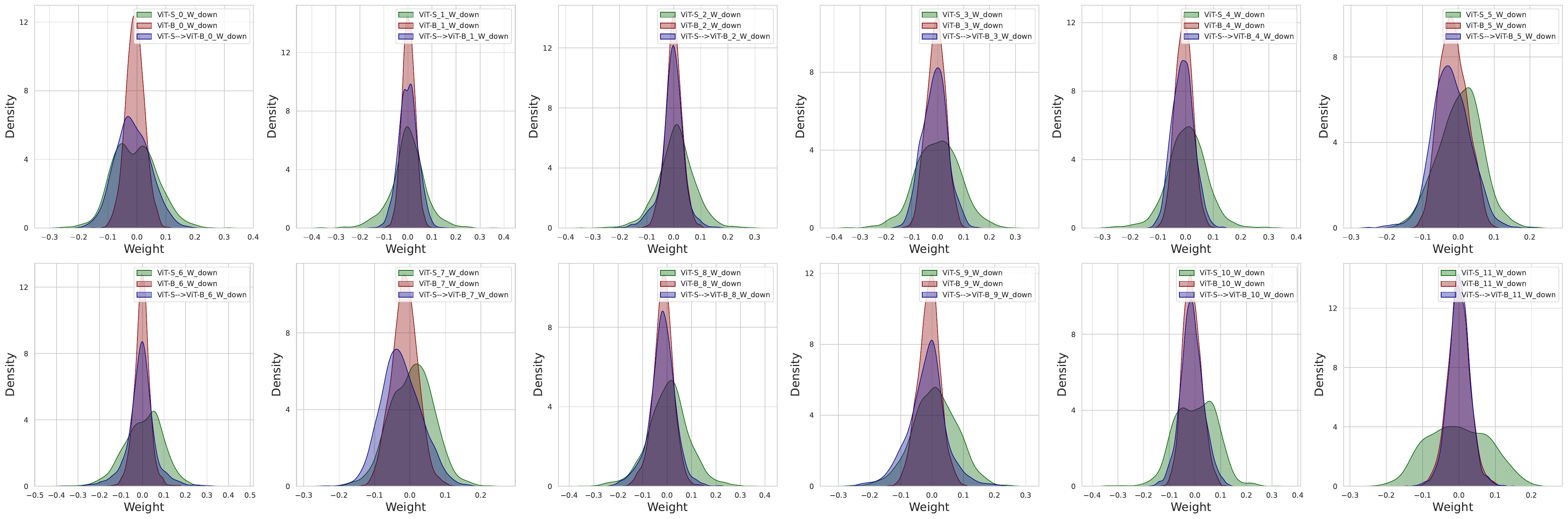}
    \caption{Visualization of the weight distribution of $W_{down}$ in the adapter module on CIFAR-100 dataset.}
    \label{fig_all_wdown_after}
\end{figure*}

\begin{figure*}
    \centering
    \includegraphics[width=0.98\textwidth]{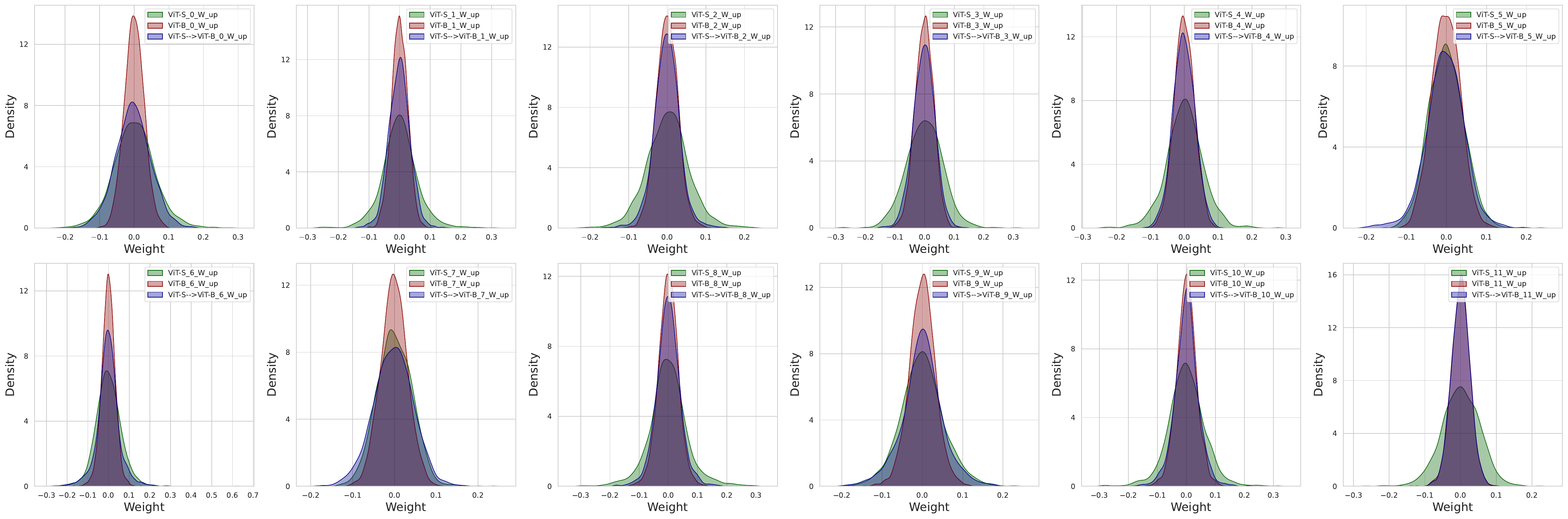}
    \caption{Visualization of the weight distribution of $W_{up}$ in the adapter module on CIFAR-100 dataset.}
    \label{fig_all_wup_after}
\end{figure*}

Using the CIFAR-100 dataset, we employ the Parallel Adapter as the AT method to fine-tune Vision Transformers at different scales: ViT-S, ViT-B, and ViT-L. We present the weight distribution of each adapter module post-fine-tuning, as illustrated in Figures \ref{fig_all_wdown_before} and \ref{fig_all_wup_before}. Given that both ViT-S and ViT-B consist of 12 layers, and ViT-L has a total of 24 layers, we visualize only the even-numbered layers for ViT-L. Subsequently, we leverage our iDAT framework with ViT-S serving as a lightweight teacher to provide distinct perspectives of knowledge for ViT-B. The post-fine-tuning weight distribution of each adapter module is depicted in Figures \ref{fig_all_wdown_after} and \ref{fig_all_wup_after}. It is discernible that, upon utilizing the iDAT framework, the weight distribution for ViT-B becomes more dispersed and less sparse, indicating that the teacher model has indeed infused new knowledge into the student model.

\section{Datasets}

In this section, we present detailed information about the VTAB-1K benchmark  \cite{vtab1k} used in this paper, as shown in Table \ref{tablevtab1k}.

\begin{table*}[!t]
    \centering
    \begin{tabular}{c|c|ccc|c}
    \hline
    \textbf{Group} & \textbf{Dataset} &\textbf{Train} & \textbf{Val} & \textbf{Test} & \textbf{\# Class} \\ \hline
    \multirow{7}{*}{\textbf{Natural}} & \textbf{CIFAR100} \cite{cifar100} &\multirow{7}{*}{800/1,000} &\multirow{7}{*}{200} & 10,000 &100  \\
         & \textbf{Caltech101} \cite{Caltech101} & & & 6,084 &102  \\
         & \textbf{DTD} \cite{dtd} & & & 1,880 &47  \\
         & \textbf{Oxford-Flowers102} \cite{flowers102} & & & 6,149 &102  \\
         & \textbf{Oxford-Pets} \cite{OxfordPets} & & & 3,669 &37  \\
         & \textbf{SVHN} \cite{SVHN} & & & 26,032 &10  \\
         & \textbf{Sun397} \cite{Sun397} & & & 21,750 &397  \\
         \hline

    \multirow{4}{*}{\textbf{Specialized}} & \textbf{Patch Camelyon} \cite{PatchCamelyon}&\multirow{4}{*}{800/1,000} &\multirow{4}{*}{200} & 32,768 &2  \\
         & \textbf{EuroSAT} \cite{Eurosat} & & & 5,400 &10  \\
         & \textbf{Resisc45} \cite{Resisc45} & & & 6,300 &45  \\
         & \textbf{Retinopathy} \cite{Retinopathy} & & & 42,670 &5  \\
         \hline

    \multirow{8}{*}{\textbf{Structured}} & \textbf{Clevr/count} \cite{Clevr} &\multirow{8}{*}{800/1,000} &\multirow{8}{*}{200} & 15,000 &8  \\
         & \textbf{Clevr/distance} \cite{Clevr} & & & 15,000 &6  \\
         & \textbf{DMLab} \cite{DMLab} & & & 22,735 &6  \\
         & \textbf{KITTI-Dist} \cite{KITTI} & & & 711 &4  \\
         & \textbf{dSprites/location} \cite{dSprites} & & & 73,728 &16  \\
         & \textbf{dSprites/orientation} \cite{dSprites} & & & 73,728 &16  \\
         & \textbf{SmallNORB/azimuth} \cite{SmallNORB} & & & 12,150 &18  \\
         & \textbf{SmallNORB/elevation} \cite{SmallNORB} & & & 12,150 &18  \\
         \hline
    \end{tabular}
    \caption{The details of the VTAB-1K benchmark.}
    \label{tablevtab1k}
\end{table*}

\section{Implementation details}

Following the settings of SSF \cite{ssf}, we directly resize the image to 224 × 224. We employ AdamW \cite{adamw} as the optimizer, set the batch size to 32, and designate 100 epochs with a provision for a 10 epochs warm-up at a warmup learning rate of 1e-7. Regarding the initial learning rate (lr), previous study \cite{ssf} have established different values for various datasets, as detailed in Table \ref{tablecvlr}.

\begin{table*}[!t]
    \centering
    \footnotesize
    \setlength{\tabcolsep}{2pt}
    \begin{tabular}{c|ccccccccccccccccccc}
    \hline
      Dataset   & \rotatebox{90}{CIFAR-100} & \rotatebox{90}{Caltech101} & \rotatebox{90}{DTD}  & \rotatebox{90}{Flowers102} & \rotatebox{90}{Pets} & \rotatebox{90}{SVHN} & \rotatebox{90}{Sun397} & \rotatebox{90}{Patch Camelyon} & \rotatebox{90}{EuroSAT} & \rotatebox{90}{Resisc45} & \rotatebox{90}{Retinopathy} & \rotatebox{90}{Clevr/count} & \rotatebox{90}{Clevr/distance} & \rotatebox{90}{DMLab} & \rotatebox{90}{KITTI/distance} & \rotatebox{90}{dSprites/loc} & \rotatebox{90}{dSprites/ori} & \rotatebox{90}{SmallNORB/azi} & \rotatebox{90}{SmallNORB/ele}  \\ \hline
      Initial lr  &5e-3 &1e-3 &5e-3 &5e-3 &5e-3 &1e-2 &5e-3 &5e-3 &3e-3 &2e-3 &5e-3 &2e-3 &5e-2 &5e-3 &1e-2 &1e-2 &5e-3 &2e-2 & 5e-3 \\
      \hline
    \end{tabular}
    \caption{The detailed initial learning rate on the VTAB-1K benchmark.}
    \label{tablecvlr}
\end{table*}

\end{document}